\documentclass{article}

\PassOptionsToPackage{numbers, compress}{natbib}

    \usepackage[preprint]{neurips_2020}

\usepackage[utf8]{inputenc} 
\usepackage[T1]{fontenc}    
\usepackage{hyperref}       
\usepackage{url}            
\usepackage{booktabs}       
\usepackage{multirow}
\usepackage{amsfonts}       
\usepackage{nicefrac}       
\usepackage{microtype}      
\usepackage[pdftex]{graphicx}
\usepackage{epstopdf}
\usepackage{algorithm}  
\usepackage{algorithmicx}  
\usepackage{algpseudocode}  
\usepackage{amsmath}  
\usepackage{subfigure}
\usepackage{wrapfig}

\usepackage{comment}
\usepackage{xcolor}

\usepackage{amsthm}
\newtheorem{definition}{Definition}
\newcommand{\tabincell}[2]{\begin{tabular}{@{}#1@{}}#2\end{tabular}}

\title{LotteryFL: Personalized and Communication-Efficient Federated Learning with Lottery Ticket Hypothesis on Non-IID Datasets}

\author{
  Ang Li  
  \\Duke University\\ 
  \texttt{ang.li630@duke.edu} \\
  \And
  Jingwei Sun \\
  Duke University \\
  \texttt{jingwei.sun@duke.edu} \\
  \And
  Binghui Wang \\
  Duke University\\
  \texttt{binghui.wang@duke.edu} \\
  \And
  Lin Duan \\
  Duke University\\
  \texttt{lin.duan@duke.edu} \\
  \And
  Sicheng Li \\
  Alibaba DAMO Academy\\
  \texttt{sicheng.li@alibaba-inc.com} \\
  \And
  Yiran Chen\\
  Duke University\\
  \texttt{yiran.chen@duke.edu}\\
  \And
  Hai Li\\
  Duke University\\
  \texttt{hai.li@duke.edu}
}

\begin{document}

\maketitle

\begin{abstract}
Federated learning is a popular distributed machine learning paradigm with enhanced privacy. Its primary goal is learning a global model that offers good performance for the participants as many as possible. 
The technology is rapidly advancing with many unsolved challenges, among which \textit{statistical heterogeneity (i.e., non-IID)} and \textit{communication efficiency} are two critical ones that hinder the development of federated learning. 
In this work, we propose LotteryFL -- a personalized and communication-efficient federated learning framework via exploiting the Lottery Ticket hypothesis. 
In LotteryFL, each client learns a lottery ticket network (i.e., a subnetwork of the base model) by applying the Lottery Ticket hypothesis, and only these lottery networks will be communicated between the server and clients.
Rather than learning a shared global model in classic federated learning, each client learns a personalized model via LotteryFL; 
the communication cost can be significantly reduced due to the compact size of lottery networks.
To support the training and evaluation of our framework, we construct non-IID datasets based on MNIST, CIFAR-10 and EMNIST by taking \textit{feature distribution skew}, \textit{label distribution skew} and \textit{quantity skew} into consideration.
Experiments on these non-IID datasets demonstrate that LotteryFL significantly outperforms existing solutions in terms of personalization and communication cost.
  
\end{abstract}

\section{Introduction}
Federated learning (FL) \cite{mcmahan2017communication} is a popular distributed machine learning framework that enables a number of clients to train a shared global model collaboratively without transferring their local data. A central server coordinates the FL process, where each participating client communicates only the model parameters with the central server while keeping local data private. In this way, FL overcomes privacy challenges and allows machine learning models to learn decentralized data. 
FL has been applied to many practical applications where data is distributed across clients and too sensitive to be aggregated into a central repository. For example, FL has demonstrated its good performance for the next-word-prediction task on smartphones \cite{hard2018federated}.

The clients that participate in the FL process expect to obtain a shared global model that can provide better performance than the models individually trained by themselves.
In practice, 
the distribution of the data across the clients is inherently non-IID (non-identically independently distributed). 
Such statistical heterogeneity makes it difficult to train a shared global model that can be well generalized for all clients~\cite{mcmahan2017communication, li2019federated}. 
Many studies attempt to mitigate the statistical heterogeneity via performing personalization in FL, including exploiting meta-learning \cite{jiang2019improving,khodak2019adaptive,chen2018federated}, multi-task learning \cite{smith2017federated,zantedeschi2019fully}, transfer learning \cite{wang2019federated,mansour2020three}, etc. 
These approaches commonly require two separate steps: 1) training a global model collaboratively and 2) adapting the model to each client by using its local data. 
Such a two-step process for personalization inevitably induces extra overheads. 
Communication cost is another major bottleneck of FL as the communication links between the central server and the participating clients typically operate at a low rate and could be expensive.
Thus, a straightforward approach to alleviating the communication bottleneck is compressing the data communicated between the server and the clients.
Common practices include sparsification \cite{konevcny2016federated}, quantization \cite{alistarh2017qsgd}, etc. 
However, very few efforts have been spent on addressing these two critical challenges simultaneously, 
The only possible exception is LG-FedAvg, which is proposed by Liang et al. \cite{liang2020think}. 
However, LG-FedAvg was built based on an unrealistic FL setting, where each client holds sufficient training data (300 images/class of MNIST and 250 images/class of CIFAR-10). 
Such a condition implies that a client has already been able to obtain a good performance by training a model locally, which 
is indeed against the motivation of FL.
Instead, in this work, we consider a more realistic and challenging scenario, where each client owns limited data (e.g., 5 images/class). 
As such, none of the clients is able to train a local model with desired performance.

\textbf{Our work:}  We design LotteryFL -- a personalized and communication-efficient FL framework via exploiting the Lottery Ticket hypothesis~\cite{frankle2018lottery}. The Lottery Ticket hypothesis offers a simple solution to discover the Lottery Ticket Networks (LTNs), which are sparse subnetworks within a large base model. 
Surprisingly, the performance of these LTNs often exceeds that of the non-sparse base model given the same training efforts. Inspired by this property, we propose to seek the LTN of each client during each communication round, and then communicate only the parameters of LTNs between the clients and the server in FL.
After aggregating the LTNs of the clients, the server will distribute the updated parameters of the corresponding LTN to each client. 
Finally, a personalized model, instead of a shared global model, will be learned at each client. Since the LTN is determined by pruning the base model using the local data of each client, the data-dependent features have already been incorporated in the LTN. 
Considering the non-IID data distribution across the clients, the LTN of each client may not significantly overlap with each other.
Therefore, the personalization property of each LTN can be retained after the aggregation is performed on the server. Additionally, due to the compact size of the LTN, the size of the model parameters that needs to be communicated is reduced. 
The communication efficiency of FL can be significantly improved accordingly.

It is worth nothing that no benchmark datasets are yet delicately designed for supporting the research on FL under non-IID settings. 
Therefore, we also construct and publish several  datasets\footnote{https://github.com/jeremy313/non-iid-dataset-for-personalized-federated-learning} to represent the characteristics of practical FL environments under non-IID settings. These non-IID datasets are constructed based on three classical datasets: MNIST \cite{lecun1995learning}, CIFAR-10 \cite{krizhevsky2009learning}, and EMNIST \cite{cohen2017emnist}. In order to quantitatively evaluate the degree of non-IID data distribution across clients, we define a metric named \textit{Client-Wise Non-IID Index (CNI)}. 
Based on \textit{CNI}, we explore the impact of non-IID distribution on our model performance. Our key contributions be can summarized as follows:
\begin{itemize}
    \item We propose a novel FL framework, namely, LotteryFL, that can achieve both personalization and high communication efficiency under non-IID setting;
    \item Based on the classical datasets, we construct several datasets to support FL under non-IID settings. We also define a metric -- \textit{Client-Wise Non-IID Index}, to quantitatively evaluate the degree of non-IID data distribution across clients. 
    \item We conduct experiments on our designed non-IID datasets and compare LotteryFL with two existing methods -- FedAvg~\cite{mcmahan2017communication} and LG-FedAvg~\cite{liang2020think}. Experimental results shows that LotteryFL significantly outperforms the compared methods in terms of both personalization and communication cost.
\end{itemize}

\section{Related Work}
FL \cite{mcmahan2017communication} is a distributed machine learning paradigm with enhanced privacy. The primary goal of FL is to learn a global model that achieves good performance for almost all participants.  FedAvg proposed by McMahan \textit{et al.}~\cite{mcmahan2017communication} is one of the most widely adopted FL methods, which adopts averaging as its aggregation method over the local models of participating clients. However, FL is still facing many challenges \cite{kairouz2019advances,li2019federated}, among which statistical heterogeneity and communication efficiency are two critical ones that hinder the development of FL. 

\vspace{-2mm}
\paragraph{Personalization}
Due to statistical heterogeneity (i.e., non-IID data distribution across clients), it is necessary to adapt the  global model to achieve personalization. Existing works achieve personalization via meta-learning \cite{jiang2019improving,khodak2019adaptive,chen2018federated}, multi-task learning \cite{smith2017federated,zantedeschi2019fully}, transfer learning \cite{wang2019federated,mansour2020three}, etc. 
However, all the existing works achieve personalization in two separate steps that are associated with extra overhead: 1) a global model is learned in a federated fashion, and 2) the global model is fine-tuned for each client using the local data.

\vspace{-2mm}
\paragraph{Communication Efficiency}
Communication is a major bottleneck for FL since the communication links between clients and the server typically operate at low rates and can be expensive. Some studies \cite{alistarh2017qsgd,konevcny2016federated,ivkin2019communication} aim to reduce communication costs in FL. The key idea is to reduce the size of the data communicated between the server and clients via combining FedAvg with data compression techniques, e.g., sparsification, quantization, sketching, etc. 

To the best of our knowledge, Liang \textit{et al.}~\cite{liang2020think} proposed the first FL method, namely LG-FedAvg, to address the above two challenges simultaneously. However, the problem setting in LG-FedAvg cannot represent the realistic FL environment.
In this paper, we develop a personalized and communication-efficient FL method under a more realistic FL settings.


\section{Design of LotteryFL}

\begin{wrapfigure}{r}{0.35\textwidth}
 \vspace{-4mm}
\centering
     \includegraphics[scale=0.3]{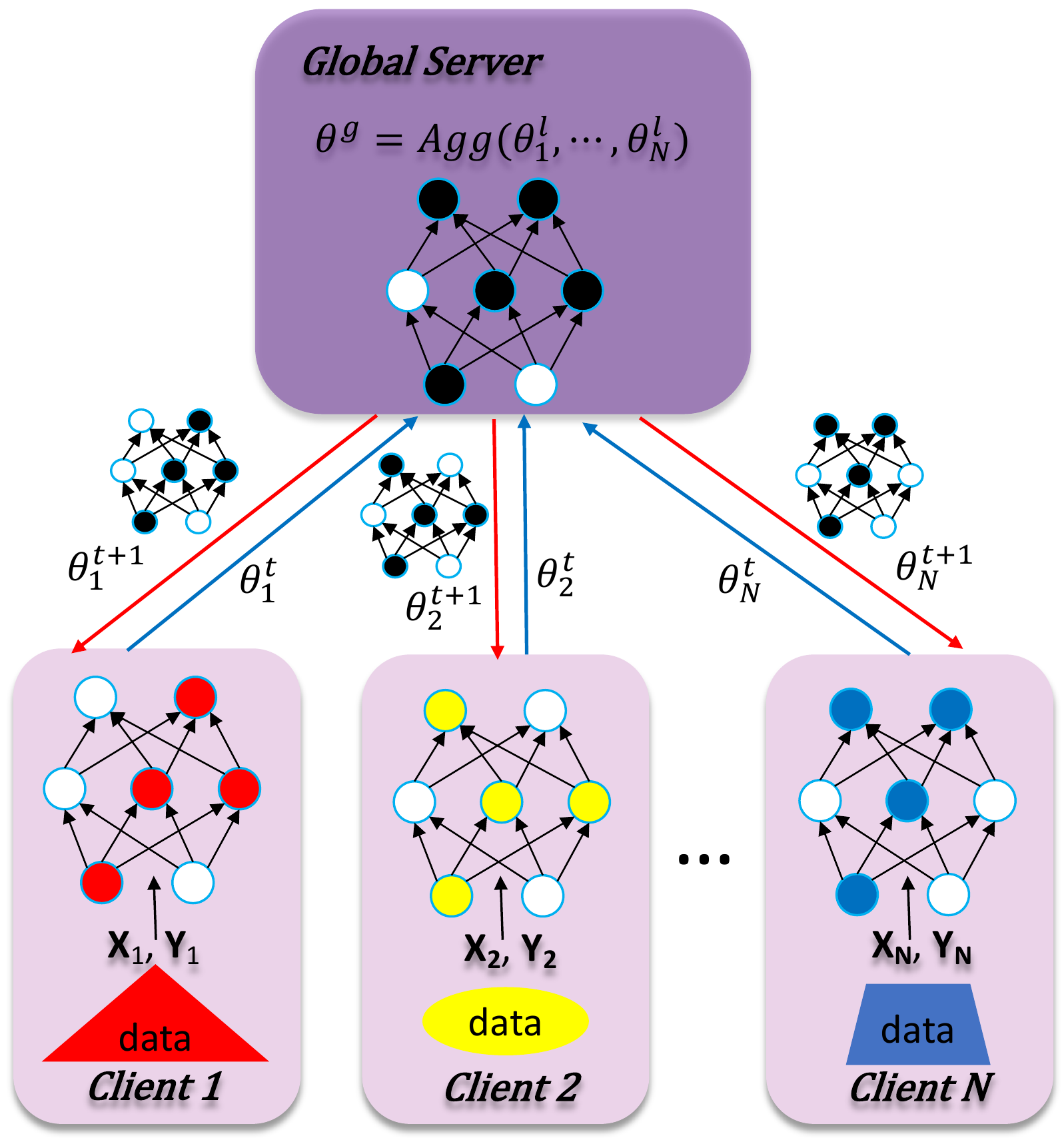}
     \vspace{-5mm}
\caption{Overview of LotteryFL.}
\label{fig:overview}
\vspace{-5mm}
\end{wrapfigure}
At high level, LotteryFL combines the Lottery Ticket hypothesis with FedAvg in an end-to-end manner. An overview of LotteryFL is shown in Figure \ref{fig:overview}. Each participating client learns to find a Lottery Ticket Network (LTN) via applying the Lottery Ticket hypothesis~\cite{frankle2018lottery}. In particular, the LTN is learned via pruning the base model using the local data of each client. Only the parameters of LTNs will be communicated between the clients and the server instead of the base model. The server then performs the aggregation over the received LTNs only, and the updated parameters of corresponding LTNs will be sent back to each client accordingly. The clients continue the training process after updating the parameters of LTNs. Before presenting the details of learning local LTNs and performing the global aggregation, we first define the following notations that are used in our paper.

\textbf{Notations:} We denote $\mathcal{C}=\{C_1, \dots, C_N\}$ as $N$ available clients, where $C_i$  denotes the $i$th client; and
$\mathcal{S} \subset \mathcal{C}$ as a set of randomly selected clients in each training round. 
Let $\theta_g$ be the parameters of the base model on the global server, and 
$\theta_i$ ($i\neq g$) represent the local model parameters on each client $C_i$. 
We also use the superscript $t$, e.g., $\theta_i^t$, to represent the model parameters learned in the $t$-th round.
Each client $C_i$ also learns a local mask $m_i\in\{0,1\}^{|\theta_i|}$, which indicates the LTN identified via applying the Lottery Ticket hypothesis. Therefore, $\theta_i\odot m_i$ denotes the parameters of the corresponding LTN at client $C_i$. 
Given the  data $D_i$ held by $C_i$, 
we split $D_i$ into the training data $D_i^{train}$, validation data $D_i^{val}$, and test data $D_i^{test}$.

\subsection{Training Algorithm}
Compared to FedAvg, the key difference is that only LTNs are communicated between the clients and the server in FL. As a result, the aggregation on the server is also performed on the LTNs only in each communication round. The details of the training algorithm for LotteryFL is elaborated in Algorithm \ref{alg:train}. In general, the training algorithm has the following steps:

\textbf{Step I:} Given the $t$th communication round, the server randomly samples a set of clients $S$.

\textbf{Step II:} Each client $C_k\subset S$ downloads its corresponding LTN $\theta_{k}^t$ from the server, where $\theta_{k}^t=\theta_g^t\odot m_k^t$. Here, $m_k^t$ is the local mask of $C_k$, which indicates the LTN inside $\theta_k^t$. 

\textbf{Step III:} Each client $C_k$ starts training the local model with  $\theta_k^t$ and evaluates $\theta_k^t$ on the validation data $D_k^{val}$. If the validation performance is better than a predefined threshold $acc_{threshold}$ and the current pruning rate $r_k^t$ does not reach the target pruning rate $r_{target}$, the client will prune the small weights of $\theta_k^t$ using a fixed pruning rate $r_p$. Once the pruning completes, the client can learn a mask $m_k^{t+1}$ to indicate the weights retained in $\theta_k$. In fact, $m_k^{t+1}$ is the learned LTN of $C_k$ for the next communication round, and the data-dependent features has already been also integrated with the learned mask. According to the workflow in the Lottery Ticket hypothesis, we re-initialize the weights of the LTN (i.e., $\theta_k^t\odot m_k^{t+1}$) as the initial values of these weights in $\theta_0$, which is used to randomly initialize $\theta_g$ at the beginning and stored on each client.

\textbf{Step IV:} Each client performs the mini-batch training for $E$ epochs using $D_k^{train}$, and then the updated $\theta_k^{t+1}$ will be sent to the server. 

\textbf{Step V:} The server performs aggregations on the LTNs only (i.e., $\theta_k^{t+1}$) via FedAvg, and updates the corresponding parameters in  $\theta_g^{t}$ to obtain $\theta_g^{t+1}$.

The above process repeats until reaches a predefined number of communication rounds. Finally, each client learns a personalized model $\theta_k$.

\vspace{-2mm}
\begin{algorithm}[t]  
\renewcommand{\algorithmicrequire}{\textbf{Server executes:}}
\renewcommand{\algorithmicensure}{\textbf{ClientUpdate($C_k$, $\theta_{k}^t$, $\theta_0$):}}
    \caption{Training Algorithm of LotteryFL.}  
    \label{alg:Framwork}  
    \begin{algorithmic}  
        \Require 
        \State initialize the global model $\theta_g$ with $\theta_0$
        \State $k \leftarrow \max(\ensuremath{N}\times K, 1)$ \Comment{$N$ available clients, random sampling rate $K$ }
        \State $S_t \leftarrow \{C_1,\dots,C_k\}$ \Comment{randomly selected $k$ participating clients indexed by $k$}
        \For{each round $t = 1, 2, \dots$}
            \For{each client $k \in S_t$ \textbf{in parallel}}
                \State $\theta_{k}^t=\theta_g^t\odot m_k^t$ \Comment{the mask of $k$th client $m_k$ indicates the corresponding LTN}
                \State $\theta_{k}^{t+1} \leftarrow \text{ClientUpdate}(C_k, \theta_{k}^t, \theta_0)$ 
            \EndFor
            \State $\theta_g^{t+1} \leftarrow$ (aggregate LTNs $\{\theta_k^{t+1}\}$)
        \EndFor
        \Ensure 
        \State $acc \leftarrow$ (evaluate $\theta_{k}^t$ with local validation data $D_k^{val}$)
        \If{$acc > acc_{threshold}$ and $r_k^t < r_{target}$} \Comment{$r_k^t$ is the current pruning rate of $k$th client's model, $r_{target}$ is the target pruning rate}
            \State $m_k^{t+1} \leftarrow$ (prune $\theta_{k}^t$ with the fixed pruning rate $r_p$ to get a new mask for the LTN)
            \State $\theta_{k}^{t+1} \leftarrow \theta_{k}^t\odot m_k^{t+1} $ (reset the masked parameters $\theta_{k}^t\odot m_k^{t+1}$ to corresponding values in $\theta_0$)
        \EndIf
        \State $\mathcal{B} \leftarrow$ (split local data $D_k^{train}$ into batches)
        \For{each local epoch $i$ from $1$ to $E$}
            \For{batch $b \in \mathcal{B}$}
              \State $\theta_k^{t+1} \leftarrow \theta_k^{t}  - \eta \nabla_{\theta_k^{t}} \ell(\theta_k^{t}; b)$ \Comment{$\eta$ is the learning rate, $\ell(\cdot)$ is the loss function }
            \EndFor
         \EndFor
        \State return $\theta_k^{t+1}$ to server
    \end{algorithmic} \label{alg:train}
\end{algorithm}  
\vspace{-1.5mm}

\section{Non-IID Datasets}\vspace{-1.5mm}
In this section, we first construct a set of non-IID datasets that are designed to capture the characteristics of FL in practice. 
Then, we define a novel metric  \emph{Client-Wise Non-IID Index} to quantitatively measure the degree of non-IID data distribution across clients.

\vspace{-4mm}
\subsection{Non-IID Dataset Generation}\label{subsec:non-iid}\vspace{-1.5mm}
We construct non-IID datasets based on classic datasets, such as MNIST \cite{lecun1995learning}, CIFAR-10 \cite{krizhevsky2009learning} and EMNIST \cite{cohen2017emnist}. 
Specifically, we primarily focus on \textit{feature distribution skew}, \textit{label distribution skew} and \textit{quantity skew}, which are three major ways that make data be non-identically distributed across clients. 

    \noindent \textit{Feature distribution skew}: It is  common that data across clients have different features, which, howver, may correspond to the same label. Handwriting recognition is a typical example of feature distribution skew, where users write the same character but with different stroke width, slant, etc.
    
    \noindent \textit{Label distribution skew}: Given a specific label, each client may hold vastly different amount of data corresponding to this label.  For example, when clients are tied to specific locations, the label distribution can be varied across clients --  certain apparel is used by one demographic but not othersla.
    
    \noindent \textit{Quantity skew}: Given the dataset on a specific client, the amount of data between different labels can be significantly unbalanced.

In practice, the real datasets for FL may contain a mixture of the above effects. To strengthen the motivation for participating the FL, we assume each client does not own sufficient data to train a local model with the desired performance. Therefore, we design two specific configurations to construct non-IID datasets based on MNIST and CIFAR-10.

    \noindent {\bf{n-class \& balanced.}} Each client holds $n$-class training data and these $n$ classes can be different across clients. The data volume of each class is balanced, however, the volume of $n$-class training data is limited such that it is infeasible to train a local model with good performance. In addition, the test data follow the same distribution as the training data. This configuration can represent the \textit{feature distribution skew} case.
    
    \noindent {\bf{n-class \& unbalanced.}} Each client holds $n$-class training data and the $n$-class can be varied across clients. The data volume of each class is unbalanced, but the amount of $n$-class data is insufficient to locally train a  model with good performance. The test data also follows the same distribution as the training data. This configuration stands for a mixture of \textit{feature distribution skew} and \textit{quantity skew}.

EMNIST, an extension of MNIST, represents a more challenging classification task involving both letters and digits. 
EMNIST shares the same image structure and parameters as the original MNIST. 
Moreover, each image in EMNIST is also associated with an attribute `By\_Author', which represents the writer of each image. Therefore, EMNIST can be naturally transformed to a non-IID dataset by grouping the handwritten characters images based on the writer. Then, each client holds only a specific writer's handwritten characters images in our experiments.

\subsection{Client-Wise Non-IID Index}\vspace{-1.5mm}
To explore the impact of non-IID data distribution on the model performance, we need to quantitatively measure the degree of non-IID across clients. He \textit{et al.}~\cite{he2019towards} define a metric named Non-IID Index (NI) to quantify the distribution shift between training data and test data. However, applying NI to different datasets requires different trained feature extractors $g_\phi(\cdot)$ and classifiers $f_\theta(\cdot)$, which are not practical in FL. 
We aim to quantify the degree of non-IID  across clients in a simple and unified way. Specifically, we propose Client-Wise Non-IID Index (CNI), which replaces the feature extractors $g_\phi(\cdot)$ in NI with a fixed encoder $En(\cdot)$.

\begin{definition}[Client-Wise Non-IID Index (CNI)]
Given an encoder $En(\cdot)$ and a participating client $C_i$ in FL, the CNI is defined as:
\begin{equation}
\small
\operatorname{CNI}(C_i)=
\left \|
\frac{\sum_{k} \frac{1}{\left|C_{i}\right|} \overline{En\left(D_{i}^k \right)}-\overline{\sum_{l} \frac{1}{\left|C_{j}\right|}_{j\neq i} \overline{En\left(D_{j}^l\right)}}}{\sigma(En(D))}\right\|_2,
\label{eq:cni}
\end{equation}
\end{definition}
where $D=\bigcup_{i=1}^N D_i$,
$D_i^k$ denotes the data belonging to the $k$th class in $D_i$.
$|C_i|$ is the number of classes in $D_i$. $\overline{(\cdot)}$ represents the first order moment, $\sigma(\cdot)$ is the standard deviation and is used to normalize the scale, and $\left\|\cdot\right\|_2$ indicates the $\ell_2$-norm.  The intuition behind Equation (\ref{eq:cni}) is to measure the distance between the average data representations from different classes in feature space on a given client and the counterpart over all the other clients. 

\section{Evaluation}\label{sec:eval}

\subsection{Experiment Setup}
In our experiments, we evaluate the performance of LotteryFL on our constructed non-IID datasets in terms of personalization and communication cost.

\begin{table}[t]
    \centering
    \caption{The CNI values on constructed non-IID datasets under different settings.}
    \resizebox{\textwidth}{!}{
        \begin{tabular}{l l || c | c | c }
            \toprule
            Dataset &  & 5 samples/class & 10 samples/class & 20 samples/class\\
            \hline
            \multirow{5}*{MNIST} & IID & 5.55 & 4.33 & 3.37\\
            \multirow{5}*{} & 2-class \& balanced & 7.38 & 6.14 & 5.59\\
            \multirow{5}*{} & 2-class \& unbalanced(0.75) & 7.69 & 6.41 & 5.74\\
            \multirow{5}*{} & 2-class \& unbalanced(0.5) & 8.01 & 6.71 & 5.89\\
            \multirow{5}*{} & 2-class \& unbalanced(0.25) & 9.31 & 7.56 & 6.10\\
            \hline
            \multirow{5}*{CIFAR10} & IID & 13.65 & 10.89 & 7.91\\
            \multirow{5}*{} & 2-class \& balanced & 15.20 & 11.70 & 9.16\\
            \multirow{5}*{} & 2-class \& unbalanced(0.75) & 16.85 & 12.43 & 9.45\\
            \multirow{5}*{} & 2-class \& unbalanced(0.5) & 18.74 & 13.51 & 10.43\\
            \multirow{5}*{} & 2-class \& unbalanced(0.25) & 22.90 & 17.14 & 12.29\\
            \bottomrule
            
        \end{tabular}}
    \label{tb:cni}
    \vspace{-6mm}
\end{table}

\vspace{-2mm}
\paragraph{Non-IID datasets}
To construct non-IID datasets based on MNIST and CIFAR-10, we transform these two datasets via applying our proposed two configurations in Section \ref{subsec:non-iid}. We respectively specify those two configurations as \textbf{2-class \& balanced} and \textbf{2-class \& unbalanced} in our experiments. In addition, to accurately quantify the degree of unbalances between the two classes on a client, we also define an \textit{balance rate} as the ratio between the data volume of one class and the counterpart of the other class. We use the parenthesized number to indicate the balance rate. For example, `unbalance(0.25)' indicates that data volume of one class is 25\% of the other class. To explore the impact of unbalances on the model performance, we vary the balance rate with \{0.25, 0.5, 0.75\} when constructing non-IID datasets. We also specify the number of samples per class on a client between $\{5, 10, 20\}$. 
We assume the MNIST and CIFAR-10 datasets are distributed across 400 clients. 
Moreover, we transform EMNIST to a non-IID dataset by grouping the handwritten characters images based on the writer. 
In particular, the EMNIST dataset is distributed across 2424 clients.

\vspace{-2mm}
\paragraph {Compared methods}
We compare LotteryFL with three  methods, i.e., \textit{FedAvg} \cite{mcmahan2017communication}, \textit{Standalone}, and \textit{LG-FedAvg} \cite{liang2020think}. \textbf{FedAvg} is a classic FL approach. In each communication round, randomly selected clients download the latest global model from the server, and then train the model using local data. Finally, the clients upload the updated models to the server and the server performs the aggregation over the updated models. \textbf{Standalone} aims to train a model locally by each client. \textbf{LG-FedAvg} is a two-step FL approach to achieve personalization, where  the global model is learned first, and then each client fine-tunes the global model using local data to achieve personalization. 

\vspace{-2mm}
\paragraph{Parameter setting}
We set the hyperparameters  $E$ = 10, $B$ = 32, $r_p$ = 0.2, and $acc_{threshold}$ = 0.5 in Algorithm \ref{alg:train}. The same settings are also applied to the compared methods if needed. We vary the target pruning rate $r_{target} \in \{0.1, 0.3, 0.5\}$. 
We randomly selected the number of  participating clients from $\{5, 10, 20 \}$ in each communication round.
The details of model architectures are presented in our supplementary material. 

\vspace{-2mm}
\paragraph{Evaluation metric} We adopt the classification accuracy of each client's test data to evaluate the performance of personalization, and report averaged accuracy over all clients. 
We use the data volume communicated between the clients and the server to measure communication costs.

\vspace{-2mm}
\subsection{Comparison of CNI on Non-IID Datasets}
We adopt VGG16\cite{simonyan2014very} pretrained on ImageNet~\cite{deng2009imagenet} as the fixed encoder $En(\cdot)$.  
Table \ref{tb:cni} reports the averaged CNI over all clients on MNIST and CIFAR-10. 
As a comparison, we also show CNI values on IID datasets, where each client holds images from all classes and each class has the same number of samples.
We observe that CNI values of non-IID datasets are much higher than those of IID datasets. Moreover, when the balance rate decreases (i.e., a higher degree of unbalance), the CNI becomes larger.

\subsection{Comparison of Personalization and Communication Efficiency}
\paragraph{Impact of the number of participating clients per round} 
In this experiment, we evaluate LotteryFL on non-IID datasets under the 2-class balanced setting, where each class consists of 20 samples. We perform the training on MNIST for 400 communication rounds, and for 2000 communication rounds on CIFAR-10 and EMNIST. Table \ref{tb:eval-clients} shows the results. 
We have the following observations: First, \emph{LotteryFL can achieve the best  personalization with  the lowest communication cost} on all the three non-IID datasets. For example, when there are 5 participating clients per round, LotteryFL(0.3) can achieve an accuracy of 89.70\% on CIFAR-10 non-IID dataset, which is 14.18\% and 43.5\% higher than that of LG-FedAvg and FedAvg, respectively. 
Morever, LG-FedAvg and FedAvg respectively consume 1.38X and 1.81X communication costs compared to LotteryFL. 
Standalone method performs the worst in most cases due to insufficient training data on each client. 
Second, if more clients participate in the training in each communication round, the personalization performance of all FL methods are improved.
The reason is that more involved clients contribute to a larger number of training data in FL.

\begin{table}[t]
\vspace{-2mm}
    \centering
    \caption{Comparison of personalization and communication cost with different number of participating clients in each communication round.}
    \resizebox{\textwidth}{!}{
        \begin{tabular}{l l || c c |c c| c c}
            \toprule
            \multirow{2}*{Dataset} & \multirow{2}*{Method} & \multicolumn{2}{c|}{5 clients} & \multicolumn{2}{c|}{10 clients} & \multicolumn{2}{c}{20 clients}\\
            \cline{3-8}
            \multirow{2}*{} & \multirow{2}*{} &\tabincell{c}{Acc\\(\%)} & \tabincell{c}{Communication\\cost (MB)} & \tabincell{c}{Acc\\(\%)} & \tabincell{c}{Communication\\cost (MB)} & \tabincell{c}{Acc\\(\%)} & \tabincell{c}{Communication\\cost (MB)}\\
            \hline
            \multirow{6}*{MNIST} & Standalone & 90.17 & 0 & 90.25 & 0 & 90.72 & 0\\
            \multirow{6}*{} & FedAvg & 96.02 & 165.94 & 96.14 & 331.89 & 96.46 & 663.76\\
            \multirow{6}*{} & LG-FedAvg & 96.74 & 131.00 & 97.76 & 262.01 & 97.87 & 524.02\\
            \multirow{6}*{} & LotteryFL(0.1) & 99.67 & {\bf{98.18}} & 99.86 & 141.18 & \bf{99.96} & {\bf{163.57}}\\
            \multirow{6}*{} & LotteryFL(0.3) & \bf{99.93} & 98.91 & \bf{99.94} & {\bf{140.30}} & 99.95 & 230.17\\
            \multirow{6}*{} & LotteryFL(0.5) & 99.86 & 103.00 & 99.93 & 165.21 & 99.96 & 304.54\\
            \hline
            \multirow{6}*{CIFAR10} & Standalone & 64.89 & 0 & 65.12 & 0 & 65.44 & 0\\
            \multirow{6}*{} & FedAvg & 46.20 & 2356.34 & 46.43 & 4712.68 & 47.67 & 9425.35\\
            \multirow{6}*{} & LG-FedAvg & 75.52 & 1793.64 & 75.89 & 3587.29 & 76.77 & 7174.58\\
            \multirow{6}*{} & LotteryFL(0.1) & 88.81 & {\bf{1142.45}} & 90.04 & {\bf{1596.57}} & \bf{90.61} & {\bf{2439.56}}\\
            \multirow{6}*{} & LotteryFL(0.3) & \bf{89.70} & 1298.19 & \bf{90.45} & 1977.41 & 89.91 & 3560.58\\
            \multirow{6}*{} & LotteryFL(0.5) & 89.68 & 1448.24 & 90.30 & 2439.93 & 90.53 & 4501.36\\
            \hline
            \multirow{6}*{EMNIST} & Standalone & 65.26 & 0 & 65.69 & 0 & 65.75 & 0\\
            \multirow{6}*{} & FedAvg & 82.08 & 20551.83 & 82.44 & 41103.67 & 83.06 & 82207.34\\
            \multirow{6}*{} & LG-FedAvg & 87.35 & 16030.43 & 87.47 & 32060.86 & 87.97 & 64121.72\\
            \multirow{6}*{} & LotteryFL(0.1) & 92.38 & {\bf{10864.41}} & 93.43 & {\bf{15273.75}} & 94.27 & {\bf{19978.86}}\\
            \multirow{6}*{} & LotteryFL(0.3) & 92.44 & 10986.62 & 93.45 & 17180.02 & 94.18 & 28351.86\\
            \multirow{6}*{} & LotteryFL(0.5) & \bf{92.56} & 11616.41 & \bf{93.54} & 20327.13 & \bf{94.90} & 37579.86\\
            \bottomrule
            
        \end{tabular}}
    \label{tb:eval-clients}
    \vspace{-2mm}
\end{table}

\begin{table}[t]
    \centering
    \caption{Comparison of personalization and communication cost with different number of samples for each class on the clients. }
    \resizebox{\textwidth}{!}{
        \begin{tabular}{l l || c c| c c| c c}
            \toprule
            \multirow{2}*{Dataset} & \multirow{2}*{Method} & \multicolumn{2}{c}{5samples/class} & \multicolumn{2}{c}{10samples/class} & \multicolumn{2}{c}{20samples/class}\\
            \cline{3-8}
            \multirow{2}*{} & \multirow{2}*{} &\tabincell{c}{Acc\\(\%)} & \tabincell{c}{Communication\\cost (MB)} & \tabincell{c}{Acc\\(\%)} & \tabincell{c}{Communication\\cost (MB)} & \tabincell{c}{Acc\\(\%)} & \tabincell{c}{Communication\\cost (MB)}\\
            \hline
            \multirow{6}*{MNIST} & Standalone & 86.84 & 0 & 89.11 & 0 & 90.72 & 0\\
            \multirow{6}*{} & FedAvg & 94.06 & 663.76 & 94.16 & 663.76 & 96.46 & 663.76\\
            \multirow{6}*{} & LG-FedAvg & 96.35 & 524.02 & 96.98 & 524.02 & 97.87 & 524.02\\
            \multirow{6}*{} & LotteryFL(0.1) & 98.95 & {\bf{173.02}} & 99.16 & {\bf{168.65}} & \bf{99.96} & {\bf{163.57}}\\
            \multirow{6}*{} & LotteryFL(0.3) & \bf{99.42} & 240.73 & \bf{99.59} & 236.92 & 99.95 & 230.17\\
            \multirow{6}*{} & LotteryFL(0.5) & 99.38 & 310.47 & 99.54 & 308.35 & 99.96 & 304.54\\
            \hline
            \multirow{6}*{CIFAR10} & Standalone & 59.55 & 0 & 64.06 & 0 & 65.44 & 0\\
            \multirow{6}*{} & FedAvg & 37.62 & 9425.35 & 43.20 & 9425.35 & 47.67 & 9425.35\\
            \multirow{6}*{} & LG-FedAvg & 70.69 & 7174.58 & 72.09 & 7174.58 & 76.77 & 7174.58\\\
            \multirow{6}*{} & LotteryFL(0.1) & \bf{85.97} & {\bf{3832.02}} & 87.31 & {\bf{3069.95}} & \bf{90.61} & {\bf{2439.56}}\\
            \multirow{6}*{} & LotteryFL(0.3) & 84.00 & 4951.63 & 87.89 & 3906.42 & 89.91 & 3560.58\\
            \multirow{6}*{} & LotteryFL(0.5) & 83.03 & 5739.96 & \bf{88.17} & 4856.26 & 90.53 & 4501.36\\
            \bottomrule
            
        \end{tabular}}
    \label{tb:eval-volume}
    \vspace{-4mm}
\end{table}

\vspace{-3mm}
\paragraph{Impact of the data volume on each client} In this experiment, we  evaluate the performance of LotteryFL with different number of samples per class on the clients. 
We keep 20 participating clients in each communication round. 
Table \ref{tb:eval-volume} shows the results. Similar to the results in Table \ref{tb:eval-clients}, we observe that LotteryFL obtains the best performance on personalization with the lowest communication cost. For example, when each client holds 20 samples per class,  LotteryFL(0.1) can reach an accuracy of 90.61\% on CIFAR-10 non-IID dataset, while the accuracy of Standalone, FedAvg, and LG-FedAvg are 65.44\%, 47.67\%, and 76.77\%, respectively. Meanwhile, FedAvg and LG-FedAvg consume 3.86X and 2.94X communication costs, respectively, compared with LotteryFL. 
Moreover, as the number of samples per class increases, the performance of LotteryFL is also improved. For example, when we increase the number of samples per class from 5 to 10, the accuracy of LotteryFL(0.5) increases from 83.03\% to 88.17\% on CIFAR-10 non-IID dataset.
Another interesting observation is that, as the number of samples per class increases, the communication cost reduces accordingly. 
For example, 
the communication cost for LotteryFL(0.1) on CIFAR-10 non-IID dataset is 3832.02Mb when there are 5 samples per class, while it reduces to 2439.56Mb when the number increases to 20. The reason is that more samples can speedup the convergence of the training, and the base model can be pruned more frequently on each client. A more compact model generates less data to be communicated with the server.

\vspace{-2mm}
\paragraph{Impact of the balance rate} Besides, we also explore the impact of the balance rate on the performance of LotteryFL. In this experiment, we keep the settings as the same as the above two experiments but specify at most 20 samples for one class on the clients. For instance, if the balance rate is 0.25, then one class contains  1-20 samples and the other one has 5 samples. The results in Table \ref{tb:eval-unbalance} also show that LotteryFL can significantly improve personalization and communication efficiency together under such challenging settings. In general, the performance of LotteryFL and the compared methods on balanced non-IID dataset is better than that under unbalanced setting. However, within all unbalanced setting, LotteryFL always outperforms the compared methods. For example, when we set balance rate to 0.25 (the worst case), LotteryFL(0.25) can reach an accuracy of 85.29\% on CIFAR-10 non-IID dataset. This number  is 34.96\%, 45.1\%, and 16.26\% higher than that is achieved by Standalone, FedAvg and LG-Fedavg, respectively.  Meanwhile, the communication cost of LotteryFL is reduced by 94\% and 48\% compared with FedAvg and LG-FedAvg, respectively.\vspace{-2.5mm}

\begin{table}[t]
\vspace{-2mm}
    \centering
    \caption{Comparison of personalization and communication cost with different balance rates.}
    \resizebox{\textwidth}{!}{
        \begin{tabular}{l l || c c| c c| c c| c c}
            \toprule
            \multirow{2}*{Dataset} & \multirow{2}*{Method} & \multicolumn{2}{c}{balanced} & \multicolumn{2}{c}{unbalanced(0.75)} & \multicolumn{2}{c}{unbalanced(0.5)} & \multicolumn{2}{c}{unbalanced(0.25)}\\
            \cline{3-10}
            \multirow{2}*{} & \multirow{2}*{} &\tabincell{c}{Acc\\(\%)} & \tabincell{c}{Communication\\cost (MB)} & \tabincell{c}{Acc\\(\%)} & \tabincell{c}{Communication\\cost (MB)} & \tabincell{c}{Acc\\(\%)} & \tabincell{c}{Communication\\cost (MB)} & \tabincell{c}{Acc\\(\%)} & \tabincell{c}{Communication\\cost (MB)}\\
            \hline
            \multirow{6}*{MNIST} & Standalone & 90.72 & 0 & 88.78 & 0 & 88.04 & 0 & 61.82 & 0\\
            \multirow{6}*{} & FedAvg & 96.46 & 663.76 & 94.52 & 663.76 & 93.71 & 663.76 & 93.13 & 663.76\\
            \multirow{6}*{} & LG-FedAvg & 97.87 & 524.02 & 97.26 & 524.02 & 95.87 & 524.02 & 95.16 & 524.02\\
            \multirow{6}*{} & LotteryFL(0.1) & \bf{99.96} & {\bf{163.57}} & 99.13 & {\bf{170.40}} & 98.75 & {\bf{175.22}} & 98.27 & {\bf{183.61}}\\
            \multirow{6}*{} & LotteryFL(0.3) & 99.95 & 230.17 & 99.40 & 240.02 & \bf{99.33} & 244.58 & 98.96 & 246.58\\
            \multirow{6}*{} & LotteryFL(0.5) & 99.96 & 304.54 & \bf{99.55} & 311.64 & 99.33 & 315.20 & \bf{99.15} & 319.03\\
            \hline
            \multirow{6}*{CIFAR10} & Standalone & 65.44 & 0 & 58.25 & 0 & 55.60 & 0 & 50.33 & 0\\
            \multirow{6}*{} & FedAvg & 47.67 & 9425.35 & 44.12 & 9425.35 & 43.04 & 9425.35 & 40.19 & 9425.35\\
            \multirow{6}*{} & LG-FedAvg & 76.77 & 7174.58 & 75.19 & 7174.58 & 72.81 & 7174.58 & 69.03 & 7174.58\\
            \multirow{6}*{} & LotteryFL(0.1) & \bf{90.61} & {\bf{2439.56}} & 88.93 & {\bf{2591.93}} & {\bf{88.53}} & {\bf{2612.29}} & 84.49 & {\bf{2973.22}}\\
            \multirow{6}*{} & LotteryFL(0.3) & 89.91 & 3560.58 & 88.71 & 3655.73 & 88.07 & 3781.99 & 85.25 & 3931.81\\
            \multirow{6}*{} & LotteryFL(0.5) & 90.53 & 4501.36 & \bf{89.40} & 4683.48 & 87.42 & 4750.81 & {\bf{85.29}} & 4848.59\\
            \bottomrule
            
        \end{tabular}}
    \label{tb:eval-unbalance}
    \vspace{-6mm}
\end{table}

\subsection{Behavior of Personalization}
\begin{wrapfigure}{r}{0.5\textwidth}
    \subfigure[MNIST]{
    \centering
    \includegraphics[scale=0.2]{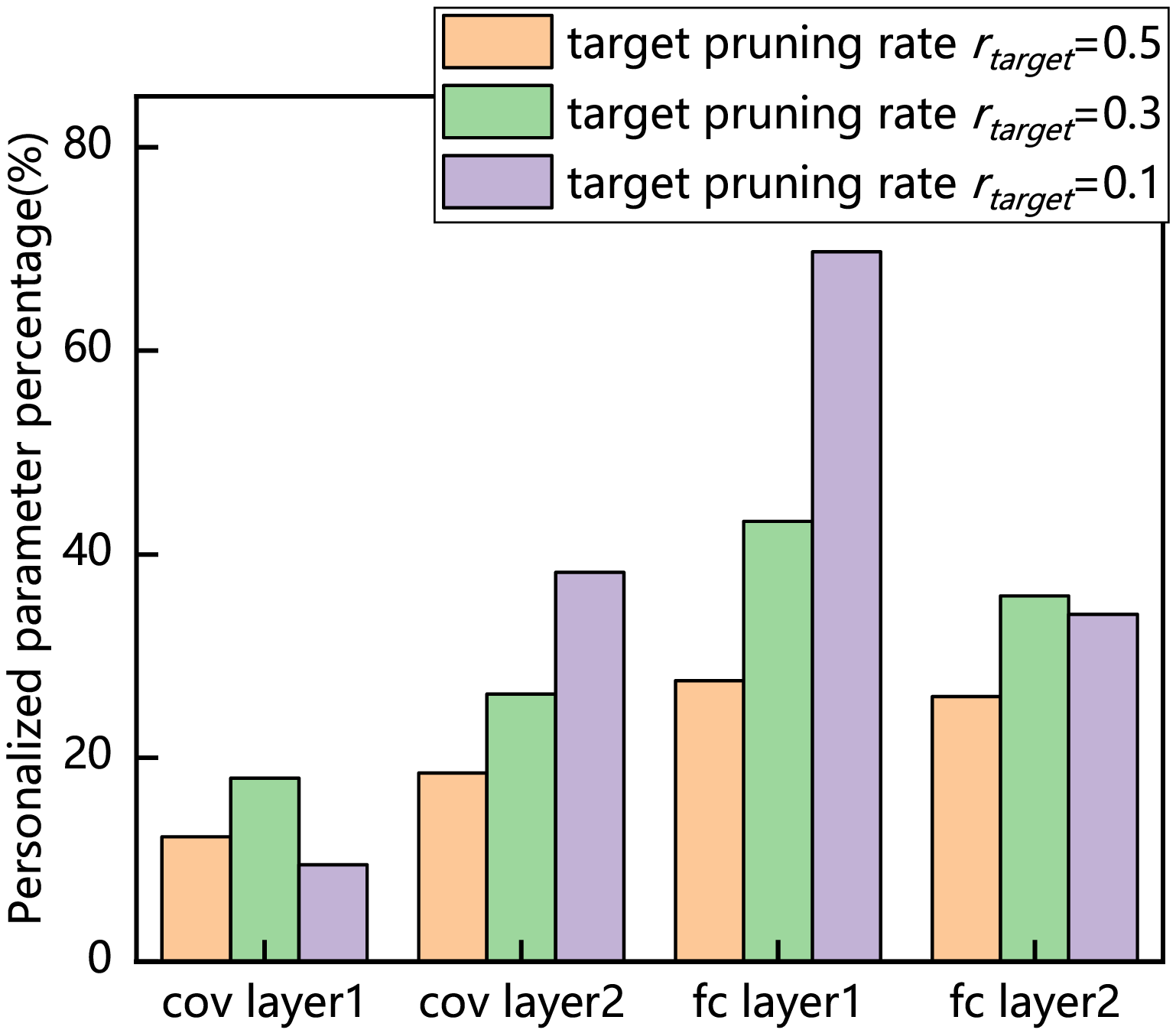}}
    \subfigure[CIFAR-10]{
    \centering
    \includegraphics[scale=0.2]{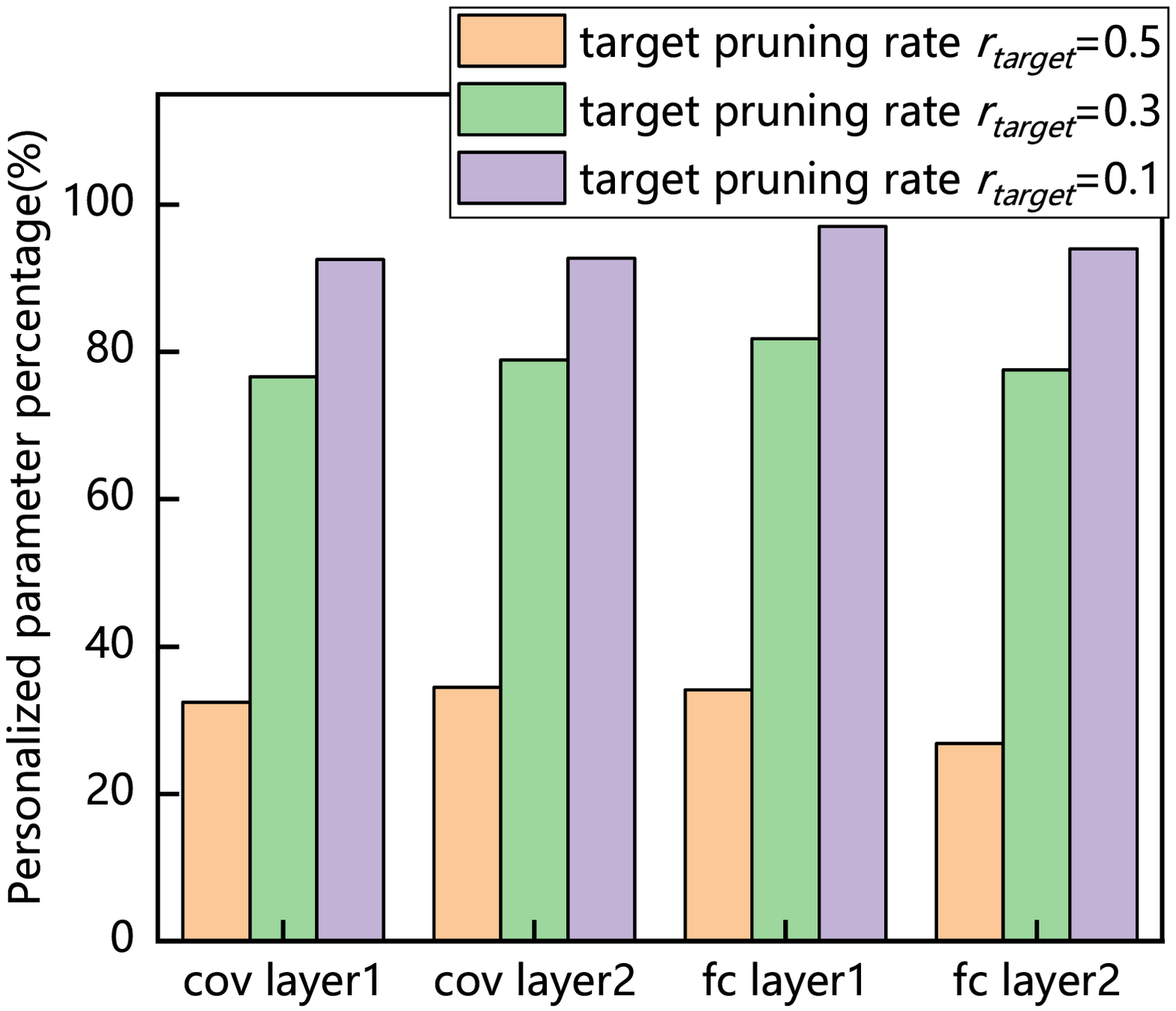}}
\caption{The distributions of personalized parameters across different layers. (`2-class balanced', 20 samples/class, 20 participating clients in each communication round)}
    \label{fig:distribution}
\end{wrapfigure}

\vspace{-2.5mm}
To better understand how LotteryFL realizes personalization, we investigate the ratio of  parameters that are unique among each client's personalized model. To this end, we define the parameters that are shared by less than 10\% of clients as the personalized parameters, and we visualize the distributions of the personalized parameters among each client's personalized model. The averaged results over all participating clients are shown in Figure \ref{fig:distribution}. 
As Figure \ref{fig:distribution} shows, with an increasing target pruning rate, the percentage of the personalized parameters of each layer will be higher. Such a phenomenon indicates that applying the iterative pruning based on the Lottery Ticket hypothesis can remove the commonly shared parameters of each layer, but retain the personalized parameters that can represent the features of local data. For example, when we set the target pruning rate to 0.1, there are as high as 93\% and 95\% personalized parameters in the first convolutional layer and the first fully connected layer of the learned models, which are trained using the CIFAR-10 non-IID dataset. This also explains why the aggregation on the server does not degrade the performance of LotteryFL, since only a small portion of the parameters are overlapped between different clients' LTNs.\vspace{-3.5mm}

\section{Conclusion}\vspace{-3.5mm}
We design LotteryFL -- a personalized and communication-efficient FL framework under non-IID settings, which is inspired by the Lottery Ticket hypothesis. We also construct and publish well-designed datasets to support FL under non-IID settings, which could facilitate the research of robust FL under more parasitical environments. In addition, we define CNI, which is the first metric to quantitatively evaluate the degree of non-IID data distribution across clients. The experimental results on non-IID datasets demonstrate that LotteryFL significantly outperforms three compared methods in terms of personalization and communication cost.

\section*{Broader Impact}
Any organizations or institutions can be considered as the clients in FL. For example, hospitals hold a huge amount of patient data for intelligent healthcare. However, under restrict privacy regulations or ethical constraints, hospitals may be required to keep the data local. FL is a promising solution for such applications, as it can enable the collaborative learning without compromising privacy. However, some works \cite{zhu2019deep,zhao2020idlg} have shown that it is feasible to recover the private training data by eavesdropping the data communicated between the clients and the server. Since only the parameters of each client's LTN will be communicated in LotteryFL and the LTN is dynamically changing during the training process, LotteryFL offers a potential solution to further enhance the privacy under such attacks.

\bibliographystyle{ieeetr}
\bibliography{reference}

\end{document}